\documentclass[runningheads]{llncs}
\usepackage[T1]{fontenc}
\usepackage{graphicx}
\usepackage{amsmath}
\usepackage{multirow} 
\usepackage{bbding}
\usepackage{makecell}
\usepackage{amssymb}
\usepackage[misc]{ifsym}
\newcommand{\corrAuthor}{$^{\textrm{\Letter}}$}
\usepackage{hyperref}
\usepackage{color}

\begin{document}
\title{\underline{Stable} Diffusion Segmentation for Biomedical Images with
Single-step Reverse Process}
\titlerunning{SDSeg: \underline{Stable} Diffusion Segmentation}
%
\author{Tianyu Lin \inst{1} \and        
Zhiguang Chen \inst{2} \and             
Zhonghao Yan \inst{3} \and \\           
Weijiang Yu \inst{2} \corrAuthor \and   
Fudan Zheng   \inst{2} \corrAuthor      
}
\authorrunning{T. Lin et al.}
%
\institute{School of Biomedical Engineering, Sun Yat-sen University, China \and 
School of Computer Science and Engineering, Sun Yat-sen University, China \\
\email{weijiangyu8@gmail.com, zhengfd3@mail2.sysu.edu.cn} \and
International School, Beijing University of Posts and Telecommunications, China 
}
\maketitle              
\begin{abstract}
Diffusion models have demonstrated their effectiveness across various generative tasks. However, when applied to medical image segmentation, these models encounter several challenges, including significant resource and time requirements. They also necessitate a multi-step reverse process and multiple samples to produce reliable predictions. To address these challenges, we introduce the first latent diffusion segmentation model, named SDSeg, built upon stable diffusion (SD). SDSeg incorporates a straightforward latent estimation strategy to facilitate a single-step reverse process and utilizes latent fusion concatenation to remove the necessity for multiple samples. Extensive experiments indicate that SDSeg surpasses existing state-of-the-art methods on five benchmark datasets featuring diverse imaging modalities. Remarkably, SDSeg is capable of generating stable predictions with a solitary reverse step and sample, epitomizing the model's stability as implied by its name.
The code is available at \url{https://github.com/lin-tianyu/Stable-Diffusion-Seg}.

\end{abstract}
\section{Introduction}
Image segmentation is a crucial task in medical image analysis. 
To alleviate the workload on medical professionals, numerous automated algorithms for medical image segmentation have been developed. The effectiveness of various neural network architectures, such as Convolutional Neural Networks (CNNs)\cite{Ronneberger2015,Chen2018} and Vision Transformers (ViT)\cite{Tang2022,hatamizadeh2022unetr}, has underscored deep learning as a promising approach to medical image segmentation.

The recent interest in Diffusion Probabilistic Models (DPM)\cite{Ho2020,Nichol2021improved} among researchers has led to a focus on image-level diffusion models in DPM-based segmentation methods\cite{Wu2023b,wu2023medsegdiffV2,Xing2023,baranchuk2022label}.
Image-level diffusion models introduce noise to an image through \textit{forward process}, and generate new images by learning how to decode this noise addition step by step in \textit{reverse process}. 
DPM-based segmentation methods utilize image conditioning to generate segmentation predictions. 
However, these approaches face limitations: 
(1) generating segmentation maps in pixel space is unnecessary and may lead to inefficient optimization and high computational costs since binary semantic maps have sparse semantic information compared to ordinary images; 
(2) diffusion models usually require multiple reverse steps to achieve detailed and varied generations, with prior diffusion segmentation models needing several samples to average for stable predictions.

To overcome these challenges, we propose a simple yet efficient segmentation framework called SDSeg, with the following contributions:
\begin{itemize}
    \item SDSeg is built on Stable Diffusion (SD)\cite{rombach2022high}, a latent diffusion model (LDM)\cite{rombach2022high,peebles2023scalable} that conducts diffusion process on a perceptually equivalent latent space with lower resolution, making the diffusion process computationally friendly. 
    \item A simple latent estimation loss is introduced to empower SDSeg to generate segmentation results on a single-step reverse process, and a concatenate latent fusion technique is proposed to eliminate the need for multiple samples.
    \item The conditioning vision encoder is set trainable to learn images' features for segmentation and adapt to multiple medical imaging domains.
    \item SDSeg performs state-of-the-art on five benchmark datasets and significantly improves diffusion-based segmentation models by reducing training resources, increasing inference speed, and enhancing generation stability.
\end{itemize}

\begin{figure}
\centering
\includegraphics[width=0.75\textwidth]{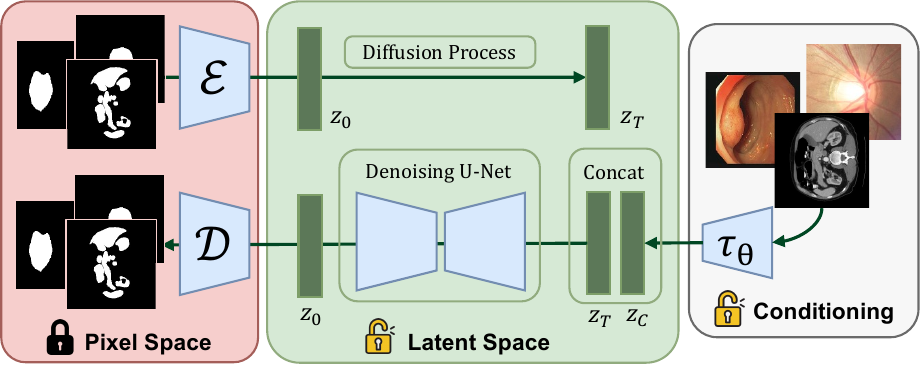}
\caption{The overview of SDSeg. We condition SDSeg via concatenation. In the training stage, we only train the denoising U-Net and vision encoder.} \label{fig:framework}
\end{figure}
\section{Methods}
The framework of SDSeg is shown in Figure.~\ref{fig:framework}. 
For medical images, we introduce a trainable vision encoder $\tau_\theta$ to encode an image $C\in \mathbb{R}^{H\times W\times 3}$ to its latent representation $z_c=\tau_\theta(C)$. 
For segmentation maps, we utilize an autoencoder for perceptual compression. As Figure.~\ref{fig:framework} shows, given a segmentation map $X\in \mathbb{R}^{H\times W\times 3}$ in pixel space, the encoder $\mathcal{E}$ encodes $X$ into a latent representation $z=\mathcal{E}(X)$, and the decoder $\mathcal{D}$ recovers the segmentation map from the latent, giving reconstructions $\widetilde{X}=\mathcal{D}(z)=\mathcal{D}(\mathcal{E}(X))$, where $z\in \mathbb{R}^{h\times w\times c}$.  In practice, we notice that the autoencoder provided by SD performs well enough for binary segmentation maps, as shown in Figure.~\ref{fig:recon-latent}. Thus, we keep the autoencoder frozen in the training stage, which makes SDSeg an end-to-end method. The diffusion process of SDSeg is conducted on the latent space.

\subsection{Latent Estimation}
For the training stage, the latent of segmentation map in the first timestep $z_0$ is added with $t$ time steps of Gaussian noise to get $z_t$. 
The forward process of the diffusion can be represented as:
\begin{equation}\label{eq1}
z_t = \sqrt{\bar{\alpha}_t}z_0+\sqrt{1-\bar{\alpha}_t}{n}
\end{equation}
where $n$ is random Gaussian noise, and $\bar{\alpha}$ is a hyperparameter for controlling the forward process. The goal of denoising U-Net in every training step is to estimate the distribution of the random Gaussian noise $n$, formulated as $\tilde{n} = f(z_t;z_c)$, where $f(\cdot)$ denotes the denoising U-Net. The noise prediction loss can be represented as $\mathcal{L}_{noise}=\mathcal{L}(\tilde{n}, n)$.

In tasks aimed at generating varied and semantically rich images, the gradual application of noise estimation in the reverse process can refine the outcomes progressively.
However, we believe that the inherently simpler segmentation maps do not substantially benefit from an extensive reverse process.
Instead, a proficiently trained denoising U-Net is capable of restoring the latent features containing all necessary structural and spatial characteristics for a segmentation map. 
Therefore, after obtaining the estimated noise $\tilde{n}$, 
we can straightforwardly derive the corresponding latent estimation through a simple transformation of Eq.~\ref{eq1}:
\begin{equation}\label{eq2}
\tilde{z}_0=\frac{1}{\sqrt{\bar{\alpha}_t}}(z_t-\sqrt{1-\bar{\alpha}_t}{\tilde{n}})
\end{equation}

This technique facilitates the addition of a supervision branch by setting the optimization goal to minimize the difference between the predicted \( \tilde{z}_0 \) and the true \( z_0 \), with the latent loss function defined as \( \mathcal{L}_{latent} = \mathcal{L}(\tilde{z}_0, z_0) \). 
Thus, the final loss function can be expressed as:
\begin{equation}
\mathcal{L}=\mathcal{L}_{noise}+\lambda\mathcal{L}_{latent}
\end{equation}
where $\lambda$ denotes the weight of the latent loss function. In practice, $\lambda$ is set to 1, and the $\mathcal{L}_{noise}$ and $\mathcal{L}_{latent}$  are mean absolute error.

It is noteworthy that utilizing \( \mathcal{L}_{noise} \) along with multiple iterations of DDIM\cite{song2020denoising} sampling can generate impressive segmentation results. The greatest contribution of introducing \( \mathcal{L}_{latent} \) lies in its ability to bypass the unnecessary reverse processes, thereby notably enhancing speed during the inference phase.

\subsection{Concatenate Latent Fusion}
\begin{figure}
    \centering
    \includegraphics[width=0.9\linewidth]{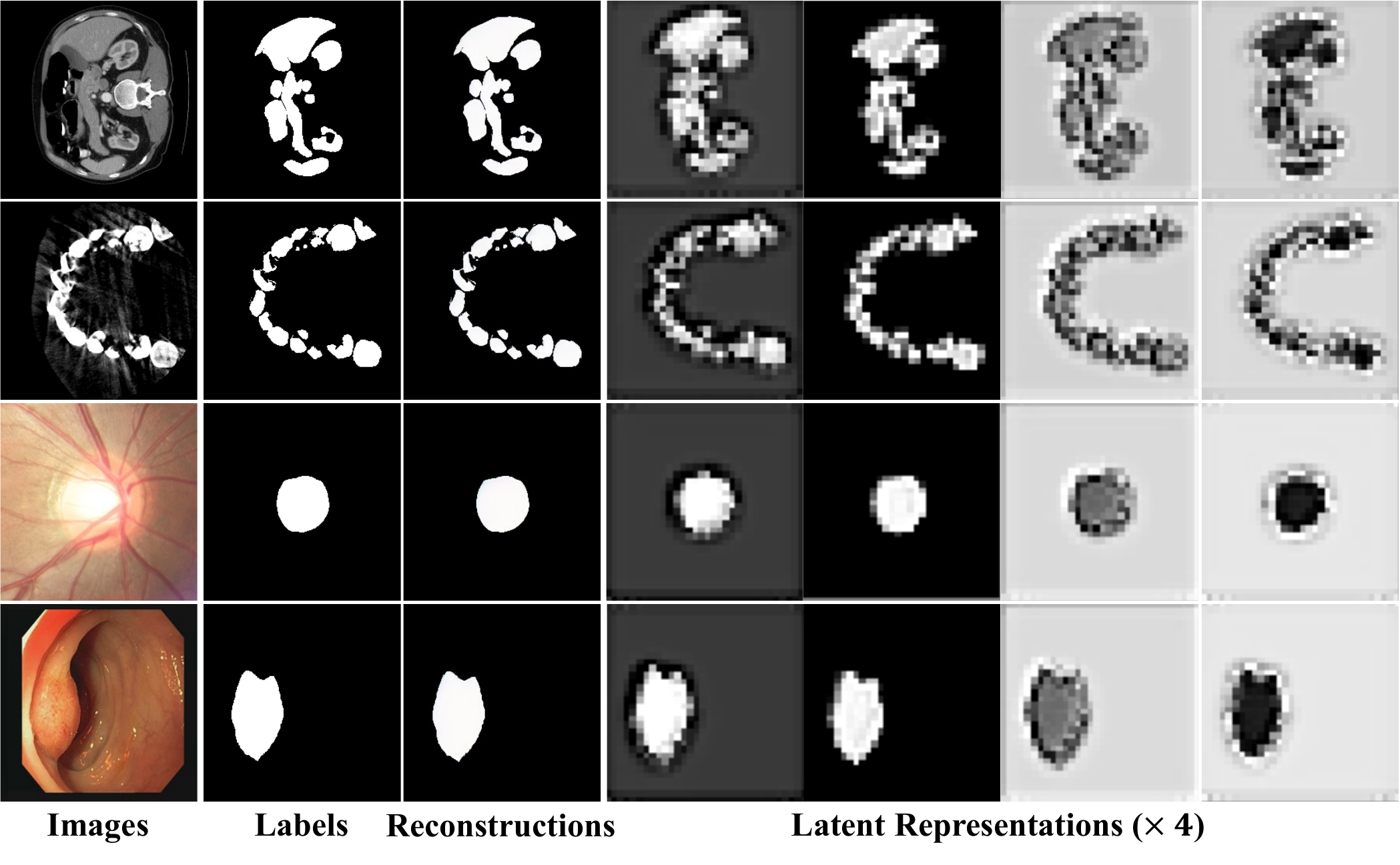}
    \caption{Visualization of reconstructions and latent representations on BTCV, STS, REF, and CVC. Reconstructions denotes $\widetilde{X}=\mathcal{D}(z)$ where latent $z=\mathcal{E}(X)$.}
    \label{fig:recon-latent}
\end{figure}
Stable Diffusion incorporates a cross-attention mechanism to facilitate multi-modal training and generation. Nonetheless, for an image-to-image segmentation model, prioritizing the extraction of semantic features and structural information from images is essential, whereas multi-modal capabilities might not offer additional advantages. Furthermore, adding cross-attention across several blocks of the denoising U-Net incurs additional computational costs. Thus, it becomes imperative to explore a more efficient method for latent fusion within SDSeg.

Moreover, our observation in Figure.~\ref{fig:recon-latent} reveals that the segmentation maps exhibit a pronounced spatial correlation with their corresponding latent representations, which might contain the necessary structural and feature information that can benefit segmentation tasks. 
Consequently, inspired by conventional semantic segmentation methods such as U-Net\cite{Ronneberger2015} and DeepLabV3+\cite{Chen2018}, etc., we employ concatenation, the prevalent and validated effective strategy for integrating an image's semantic features, to merge the latent representations of segmentation maps with those of image slices.

\subsection{Trainable Vision Encoder}
In semantic segmentation, a valid vision encoder can extract the necessary structural and semantic features from images, thereby enhancing segmentation results. As an image-conditioned generative model, SDSeg employs a trainable vision encoder to capture the abundant semantic features across images. 

The vision encoder $\tau_\theta$ has the same architecture as the encoder $\mathcal{E}$ and is initialized with its pre-trained weights.
Although we discover that simply using a frozen image encoder that is pre-trained on natural images can bring considerable results, we make the vision encoder trainable, thus allowing SDSeg to adjust to various medical image dataset modalities, enhancing its versatility and effectiveness.

\section{Experimental Results}
\subsection{Datasets and Evaluation Metrics}
To comprehensively evaluate the effectiveness and generalization ability of SDSeg, we conduct experiments on three RGB datasets on 2D segmentation task, and two CT datasets on 3D segmentation, as shown in Table.~\ref{tab:datasets}.
\begin{table}
\centering
\caption{Dataset settings. }
\label{tab:datasets}
\begin{tabular}{c|c|c|cc}
\hline
Task                                    & Dataset       &Target& Training Data & Test Data \\ \hline
\multirow{3}{*}{\makecell{2D Binary\\ Segmentation}}& CVC-ClinicDB\cite{Bernal2015} (CVC)&Polyp&      488 images         &    62 images       \\
                                        & Kvasir-SEG\cite{jha2020kvasir} (KSEG)&Polyp&     800 images         &     100 images     \\
                                        & REFUGE2\cite{Li2020,Orlando2020} (REF)&Optic Cup&     800 images         &     400 images\\ \hline
\multirow{2}{*}{\makecell{3D Binary\\ Segmentation}}& BTCV\footnotemark[1]\cite{landman2015miccai}&Abdomen Organ&      18 volumes         &     12 volumes      \\
                                        & STS-3D\cite{Cui2022,Cui2022a} (STS)&Teeth&               9 columes&           3 volumes\\ \hline
\end{tabular}
\end{table}
\footnotetext[1]{We treat all 13 organs in BTCV as a single target.}

Our evaluation encompasses three main aspects: Firstly, segmentation results across datasets are assessed using the Dice Coefficient (DC) and Intersection over Union (IoU) metrics. Secondly, we benchmark our model's efficiency by comparing its computational resource usage and inference speed against other diffusion-based segmentation methods. Thirdly, we evaluate the stability of our generated segmentation results against other diffusion segmentation models using LPIPS\cite{zhang2018unreasonable}, PSNR, SSIM, and MS-SSIM. Additionally, we conduct an ablation study to validate the efficacy of our proposed modules.

\subsection{Implementation Details}
\subsubsection{Experimental Settings} SDSeg is trained on a single V100 GPU with 16GB RAM. The model is trained for 100,000 steps using AdamW optimizer with a base learning rate of $1\times 10^{-5}$. The batch size is set to 4 by default. 
We use a KL-regularized autoencoder and LDM model with the downsampling rate $r=\frac{H}{h}=\frac{W}{w}=8$. SDSeg takes RGB images\footnotemark[2] as pixel space inputs with $H=W=256$, and the corresponding latent representation has a shape of $h=w=32$ with $c=4$. All model parts are initialized with the pre-trained weights provided by stable diffusion. The additional model parameters of the denoising U-Net for concatenate input are initialized to zeros. 
\footnotetext[2]{For 1-channel CT slices, we simply repeat 3 times to get 3-channel images.}

\subsubsection{Inference Stage} 
During the inference stage, we concatenate randomly generated Gaussian noise with the medical image's latent representation. The denoising U-Net then predicts the estimated noise, allowing SDSeg to derive the latent estimation using Eq.~\ref{eq2}. Then, Decoder $\mathcal{D}$ will transfer latent estimation to pixel space to get the final prediction. As shown in Table.~\ref{tab:resource-reverse}, SDSeg doesn’t need an external sampler and only needs a single-step reverse to sample one time for a stable prediction.

\subsection{Main Results}
\subsubsection{Comparison with State-of-the-Arts} The comparison of our model with several semantic segmentation methods on REF, BTCV, and STS datasets is shown in Table.~\ref{tab:seg}. 
We also compare our model with state-of-the-art diffusion based segmentation models on CVC, KSEG, and REF datasets, as shown in Table.~\ref{tab:sota}.  
SDSeg outperforms all other models on the five datasets with various imaging modalities, validating its effectiveness and generalization capability.
\begin{table}
    \begin{minipage}{0.48\linewidth}
        \centering
        \caption{Comparison with semantic segmentation methods, evaluated by the Dice coefficient metric. }
        \label{tab:seg}
        \begin{tabular}{c|c|c|c}
        \hline
        Methods           & REF & BTCV  & STS \\ \hline
        U-Net\cite{Ronneberger2015}              & 80.1     & 75.9& 85.4\\
        \makecell{U-Net(w/ R50)\footnotemark[3]} &          87.2& 90.5& 88.4\\
        Swin-UNETR\cite{Tang2022}        & -        & 91.3&        88.3\\
        nnU-Net\cite{Isensee2021}            & -        & 91.4&        88.9\\
        TransU-Net\cite{Chen2021a}         & 85.6     & 89.1& 88.1\\
        SwinU-Net\cite{Cao2021}          &          84.3& 86.5& 85.8\\ \hline
        Ours              & \textbf{89.4}& \textbf{92.8}& \textbf{89.4}\\ \hline
        \end{tabular}
    \end{minipage}
    \begin{minipage}{0.48\linewidth}
        \centering
        \caption{Comparison with state-of-the-art methods on REF, CVC, and KSEG.}
        \label{tab:sota}
        \begin{tabular}{c|c|cc}
        \hline
        Dataset                       & Methods               & Dice & IoU  \\ \hline
        \multirow{4}{*}{CVC} & SSFormer\cite{Wang2022c}              & 94.4 & 89.9 \\
                                      & Li-SegPNet\cite{Sharma2023}            & 92.5 & 86.0 \\
                                      & Diff-Trans\cite{chowdary2023diffusion} & 95.4 & 92.0 \\ \cline{2-4} 
                                      & Ours                  &      \textbf{95.8}&      \textbf{92.6}\\ \hline
        \multirow{4}{*}{KSEG}         & SSFormer\cite{Wang2022c}              & 93.5 & 89.0 \\
                                      & Li-SegPNet\cite{Sharma2023}            & 90.5 & 82.8 \\
                                      & Diff-Trans\cite{chowdary2023diffusion} & 94.6 & 91.6 \\ \cline{2-4} 
                                      & Ours                  &      \textbf{94.9}&      \textbf{92.1}\\ \hline
        \multirow{4}{*}{REF}& MedSegDiff- V1\cite{Wu2023b}& 86.3 & 78.2 \\
                                      & MedSegDiff-V2\cite{wu2023medsegdiffV2}         & 85.9 & 79.6 \\
                                      & Diff-Trans\cite{chowdary2023diffusion} & 88.7 & 81.5 \\ \cline{2-4} 
                                      & Ours                  &      \textbf{89.4}&      \textbf{81.8}\\ \hline
        \end{tabular}
    \end{minipage}
\end{table}
\footnotetext[3]{Replace the encoder of U-Net to a pre-trained ResNet50.}

\subsubsection{Comparison of computing resource and time efficiency} 
Table.~\ref{tab:resource-reverse} demonstrates the efficiency evaluation results of MedSegDiffs, Diff-U-Net, and SDSeg on BTCV dataset. For a fair comparison, these models are trained on the same server using their source codes. 
The results highlight SDSeg's superior efficiency, requiring significantly fewer resources and less time for training. Remarkably, SDSeg's inference process is about 100 times faster than that of MedSegDiffs and approximately 28 times quicker in generating a single segmentation map.

Table.~\ref{tab:resource-reverse} also compares the reverse process of these models. The latent estimation scheme empowers SDSeg to generate segmentation maps in a single step, and the concatenate latent fusion module allows SDSeg to sample only one time without harming model performance. Moreover, latent estimation makes SDSeg no longer rely on any external sampler for sampling.
\begin{table}
\caption{Comparison of training resources, inference speed, and reverse process settings on BTCV (1568 slices). The reverse process is assessed by $steps\times samples$.}
\label{tab:resource-reverse}
\begin{tabular}{c|cc|cc|cc|c}
\hline
Methods       & \makecell{Training\\ Time\\(hours)} & \makecell{Training\\ Resources\\($\times$ GPUs)} & \makecell{Inference\\ Time\\(hours)} & \makecell{Inference\\ Speed\\(samples/s) }  & \makecell{Diffusion\\ Sampler} &\makecell{Reverse\\ Process} & Dice \\ \hline
MedSegDiff-V1 & $\approx$ 48                        & 16GB $\times 4$                                  & $\approx$ 7                          & 0.30                                        & DPM-Solver                     &50$\times$25                 &                         79.24\\ 
MedSegDiff-V2 & {$\approx$ 49}       & {16GB $\times 4$}                 & $\approx$ 7                          & 0.31                                        & DPM-Solver                     &50$\times$25                 &                         83.52\\ 
Diff-U-Net\cite{Xing2023}\footnotemark[4]     & $\approx$  16                       & 24 GB $\times 4$                                 & $\approx$ 1/2                        & 0.87& DDIM                           &10$\times$1                  & 91.89\\ \hline
Ours          & $\approx$ 12& 16GB $\times 1$                                  & $\approx$ 1/4& 2.01& DDIM                           &10 $\times 1$                &                         92.09\\
\textbf{Ours} & \textbf{$\approx$ 12}& \textbf{16GB $\times 1$}                         & \textbf{$\approx$ 1/13}& \textbf{8.36}& \textbf{\XSolidBrush}          &\textbf{1}$\times$\textbf{1}                   & \textbf{92.76}                   \\ \hline
\end{tabular}
\end{table}
\footnotetext[4]{Diff-U-Net uses 3D sliding window infer. Inference speed is estimated as $\frac{slices}{time}$.}

\subsubsection{Stability Evaluation} Since diffusion models are generative models, the samples they generate can exhibit variability.  
However, diversity is not considered an advantageous trait in the context of medical segmentation models, as medical professionals necessitate the assistance of artificial intelligence to be consistent and reliable.
Given a trained model and fixed test data, we evaluate the stability of the diffusion-based segmentation models on the following two tasks: 
\begin{enumerate}
    \item \textbf{Dataset-level Stability}: 
    performs repeated inferences on test data to measure variability across different inferences using the LIPIS\cite{zhang2018unreasonable} metric;
    \item \textbf{Instance-level Stability}: 
    examines the model's consistency under varying initial noise by conducting repeated inferences under fixed conditions, with PSNR, SSIM, and MS-SSIM as metrics.
\end{enumerate}

Table.~\ref{tab:stability} showcases SDSeg's significant stability across these tests, underscoring its reliability in segmentation despite different initial noises.
\begin{table}
\centering
\caption{Comparison of stability evaluation on BTCV. `Seg' denotes segmentation maps; `Score' represents predicted probability scores.}
\label{tab:stability}
\begin{tabular}{c|cc|cc|cc|cc}
\hline
\multirow{2}{*}{Methods} & \multicolumn{2}{c|}{LPIPS$\downarrow$} & \multicolumn{2}{c|}{PSNR$\uparrow$}   & \multicolumn{2}{c|}{SSIM$\uparrow$}  & \multicolumn{2}{c}{MS-SSIM$\uparrow$} \\ \cline{2-9} 
                         & \multicolumn{1}{c}{Seg}  & Score  & \multicolumn{1}{c}{Seg} & Score  & \multicolumn{1}{c}{Seg} & Score & \multicolumn{1}{c}{Seg}  & Score \\ \hline
MedSegDiff-V2            & 0.3139                       & 0.2904  & 11.9271                     & 14.4506 & 0.5780                      & 0.4662 & 0.6399                       & 0.6228 \\
Diff-U-Net                & 0.0633                       & 0.0672  & 23.7158                     & 24.6675 & 0.9668                      & 0.9666& 0.9442                       & 0.9397 \\
Ours                     & \textbf{0.0199}& \textbf{0.0143}& \textbf{27.6348}& \textbf{31.5537}& \textbf{0.9796}& \textbf{0.9764}& \textbf{0.9897}& \textbf{0.9909}\\ \hline
\end{tabular}
\end{table}

\subsubsection{Ablation Study} 
\begin{figure}
    \centering
    \includegraphics[width=0.92\textwidth]{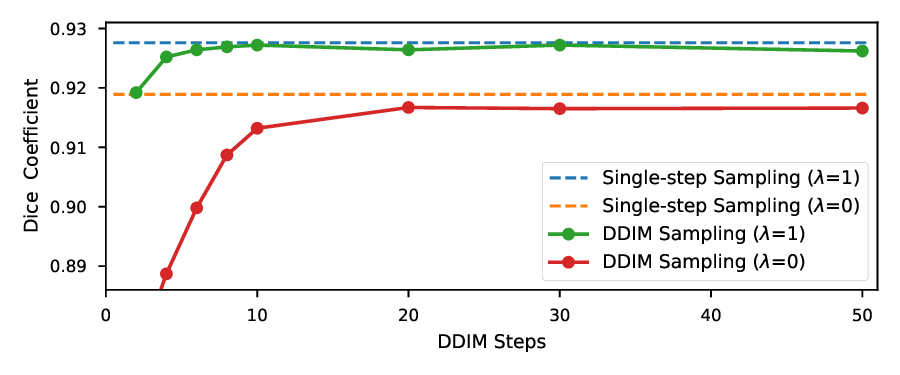}
    \caption{Comparison of DDIM convergence speed with and without latent estimation loss on BTCV. $\lambda=1$ denotes that SDSeg is trained on latent estimation loss $\mathcal{L}_{latent}$.}
    \label{fig:reverse-curve}
\end{figure}

Our ablation studies assess the contribution of each component within SDSeg, as detailed in Table.~\ref{tab:ablation}. The baseline model relies on stable diffusion with cross-attention for generating image-conditioned segmentation maps. The incorporation of latent fusion concatenation notably enhances performance, allowing for efficient learning of spatial information and features. Additionally, the trainable encoder markedly improves performances by extracting relevant semantic features from segmentation targets. While the latent estimation loss function marginally boosts performance, its primary advantage lies in significantly accelerating the reverse process, thus enabling SDSeg to discard traditional samplers for a single-step reverse process, as illustrated in Figure.~\ref{fig:reverse-curve}.
\begin{table}
\centering
\caption{Ablation study on BTCV and REF.}
\label{tab:ablation}
\begin{tabular}{c|c|c|cc|cc}
\hline
\multirow{2}{*}{Latent Estimation} & \multirow{2}{*}{\makecell{Concatenate\\ Latent Fusion}} & \multirow{2}{*}{\makecell{Trainable\\ Image Encoder}} & \multicolumn{2}{c|}{BTCV} & \multicolumn{2}{c}{REFUGE2} \\ \cline{4-7} 
                                   &                                            &                                        & Dice          & IoU          & Dice          & IoU          \\ \hline
             \XSolidBrush                      &                        \XSolidBrush                    &                        \XSolidBrush                &               32.67&              23.69&               28.31&              20.36\\
             \XSolidBrush                      &                        \CheckmarkBold                    &                        \XSolidBrush                &               80.31&              72.27&               76.79&              69.37\\
             \XSolidBrush                      &                        \CheckmarkBold                    &                        \CheckmarkBold                &               91.89&              85.41&               88.79&              80.29\\
            \CheckmarkBold                       &                       \CheckmarkBold                     &                        \CheckmarkBold                &               \textbf{92.76}&              \textbf{85.49}&               \textbf{89.36}&              \textbf{81.68}\\ \hline
\end{tabular}
\end{table}

\section{Conclusion}
In this paper, we propose SDSeg, a novel and efficient framework for medical image segmentation utilizing stable diffusion. We introduce a latent estimation strategy enabling single-step latent prediction, thereby eliminating the need for a multi-step reverse process. The model employs concatenate latent fusion for integrating learned image latent that effectively guides the segmentation task. Furthermore, a trainable vision encoder enhances the model's capability to learn image features and adapt to diverse image modalities. SDSeg achieves state-of-the-art performance across five segmentation datasets, substantially reducing training resource requirements and accelerating the inference process while maintaining remarkable stability.

\begin{credits}
\subsubsection{\ackname} 
This study was funded by 
the Program of Science and Technology of Guangdong (Grant No. 2020B1111170009), 
the 2022 Industrial Technology Basic Public Service Platform Project of China (Grant No. 2022-228-219), and
the Fundamental Research Funds for the Central Universities, Sun Yat-sen University (Grant No. 23xkjc016).

\subsubsection{\discintname}
The authors have no competing interests to declare that are
relevant to the content of this article.
\end{credits}
%
%
%
\bibliographystyle{splncs04}
\bibliography{Paper-1007}
%





\addtolength{\textheight}{11cm} 
\newpage
\appendix
\section{Stability Evaluation}
\begin{figure}
    \centering
    \includegraphics[width=1\linewidth]{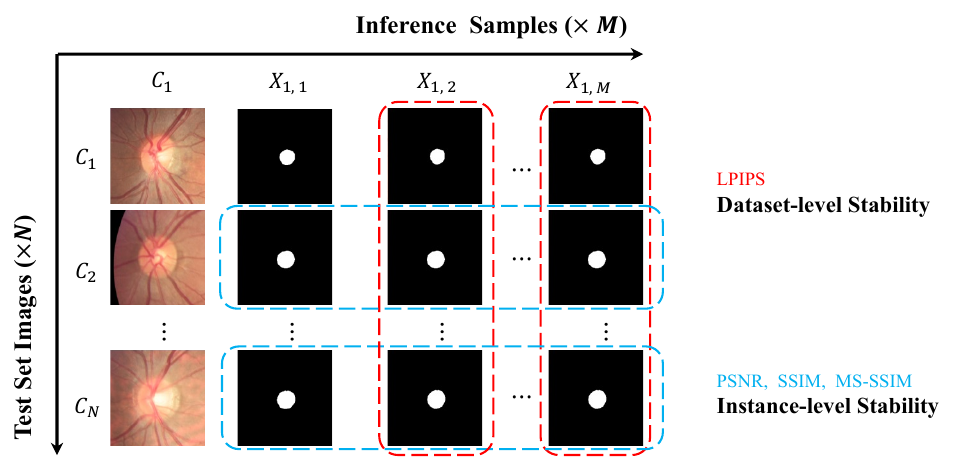}
    \caption{Illustration of our Stability Evaluation on REF. We first conduct $M$ times of inference process to prepare for the evaluation. Then, \textbf{Dataset-level Stability} is evaluated on every two sets of the inference results; \textbf{Instance-level Stability} is estimated on every two segmentation maps of each image conditioning.}
\end{figure}
\section{Qualitative Analysis}
\begin{figure}
    \centering
    \includegraphics[width=1\linewidth]{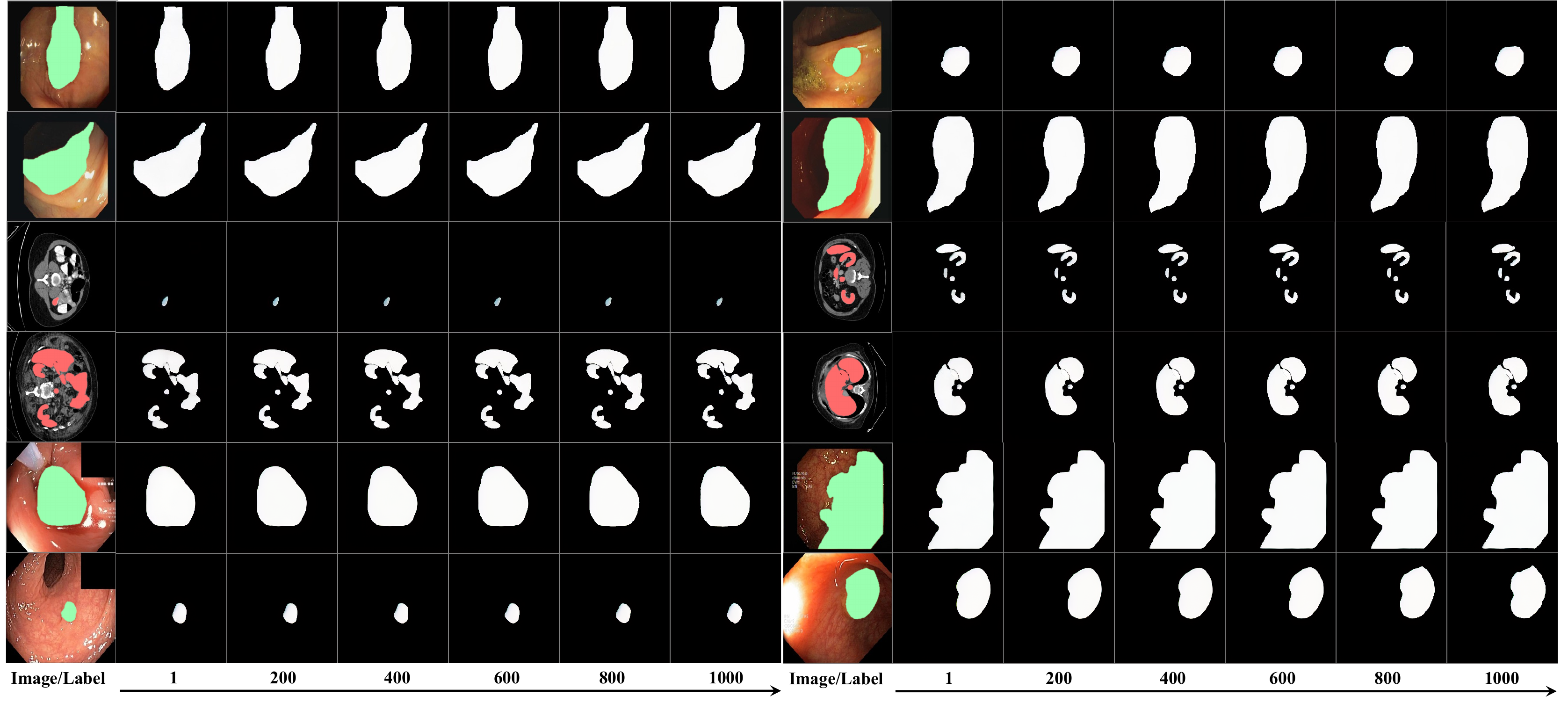}
    \caption{\textbf{From top to bottom:} Visualization of the predicted probability maps in reverse process on CVC, BTCV, and KSEG (SDSeg trained for 50,000 steps). The horizontal axis denotes DDIM sampling steps. DDIM sampler generates fine and stable results during the entire reverse process. This demonstrates that SDSeg can generate great results under limited steps of the reverse process.}
\end{figure}

\begin{figure}
    \centering
    \includegraphics[width=0.9\linewidth]{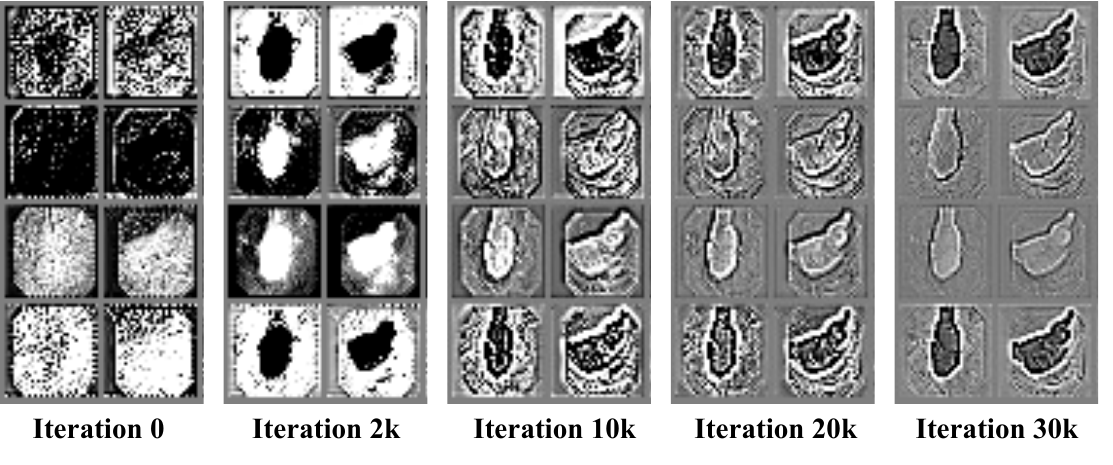}
    \caption{Visualization of the latent representations of medical images from the trainable vision encoder, on CVC. At iteration 0, the encoder pre-trained on natural images couldn't capture enough meaningful semantic features for segmentation. During training, the conditioning encoder gradually learns to focus on segmentation targets.}
    \label{fig:conditioner}
\end{figure}

\newpage
\section{The architecture of the trainable vision encoder}
We use a KL-regularized autoencoder model with the downsampling rate $r=\frac{H}{h}=\frac{W}{w}=8$. The proposed trainable vision encoder has the same network architecture as the autoencoder model's encoder.
Specifically, the trainable vision encoder's architecture can be separated into three blocks: the Downsampling block (Table.~\ref{tab:b1}), the Mid block (\ref{tab:b2}) and the Out block (\ref{tab:b3}). 

In Table.~\ref{tab:b1}, `Conv 3$\times$3' denotes convolution block with kernel size 3, `ResBlock' represents the building block in ResNet, and `Down' corresponds to downsampling. In Table.~\ref{tab:b2}, `Attention' denotes self-attention block.

\begin{table}
    \begin{minipage}{0.38\linewidth}
        \centering
        \caption{The architecture of the \textbf{Downsampling} block.}
        \label{tab:b1}
        \renewcommand\arraystretch{1.3}
        \begin{tabular}{lc}
            \hline
            \textbf{input} & $\mathbb{R}^{H \times W \times 3}$ \\ \hline
            Conv 3$\times$3 & $\mathbb{R}^{H \times W \times C}$ \\
            ResBlock$\times 2$+Down & $\mathbb{R}^{\frac{H}{2} \times \frac{W}{2} \times C}$ \\
            ResBlock$\times 2$+Down & $\mathbb{R}^{\frac{H}{4} \times \frac{W}{4} \times 2C}$ \\
            ResBlock$\times 2$+Down & $\mathbb{R}^{\frac{H}{8} \times \frac{W}{8} \times 4C}$ \\
            ResBlock$\times 2$ & $\mathbb{R}^{\frac{H}{8} \times \frac{W}{8} \times 4C}$ \\ \hline
        \end{tabular}
    \end{minipage}
    \begin{minipage}{0.28\linewidth}
        \centering
        \caption{The architecture of the \textbf{Mid} block.}
        \label{tab:b2}
        \renewcommand\arraystretch{1.5}
        \begin{tabular}{lc}
            \hline
            \textbf{input} & $\mathbb{R}^{\frac{H}{8} \times \frac{W}{8} \times 4C}$ \\ \hline
            ResBlock & $\mathbb{R}^{\frac{H}{8} \times \frac{W}{8} \times 4C}$ \\
            Attention & $\mathbb{R}^{\frac{H}{8} \times \frac{W}{8} \times 4C}$ \\
            ResBlock & $\mathbb{R}^{\frac{H}{8} \times \frac{W}{8} \times 4C}$ \\ \hline
        \end{tabular}
    \end{minipage}
    \begin{minipage}{0.28\linewidth}
        \centering
        \caption{The architecture of the \textbf{Out} block.}
        \label{tab:b3}
        \renewcommand\arraystretch{1.5}
        \begin{tabular}{lc}
            \hline
            \textbf{input} & $\mathbb{R}^{\frac{H}{8} \times \frac{W}{8} \times 4C}$ \\ \hline
            GroupNorm & $\mathbb{R}^{\frac{H}{8} \times \frac{W}{8} \times 4C}$ \\
            Conv 3$\times$3 & $\mathbb{R}^{\frac{H}{8} \times \frac{W}{8} \times 2Z}$ \\ 
            Conv 1$\times$1 & $\mathbb{R}^{\frac{H}{8} \times \frac{W}{8} \times Z}$ \\ \hline
        \end{tabular}
    \end{minipage}
\end{table}

The input segmentation map $X\in\mathbb{R}^{H \times W \times 3}$ successively goes through these three blocks to get its corresponding latent representation $z\in\mathbb{R}^{\frac{H}{8} \times \frac{W}{8} \times Z}$, where $C=128$ is the channel dimension of the vision encoder, and $Z=4$ is the channel dimension of the latent representation.
\end{document}